\documentclass[a4paper]{article}

\usepackage{INTERSPEECH2021}

\usepackage{graphicx}
\graphicspath{{./}{Figures/}}

\usepackage{xcolor}
\definecolor{HighlightColor}{rgb}{0.69,0.81,0.11} %
\definecolor{HighlightColor1}{rgb}{0.462745098, 0.725490196, 0.000000000}
\definecolor{HighlightColor2}{rgb}{0.945, 0.349, 0.373}
\definecolor{HighlightColor3}{rgb}{0.349, 0.604, 0.827}
\definecolor{HighlightColor4}{rgb}{0.976, 0.651, 0.353}
\definecolor{HighlightColor5}{rgb}{.62, 0.4, 0.671}
\definecolor{HighlightColor6}{rgb}{.804, 0.439, 0.345}
\definecolor{HighlightColor7}{rgb}{.843, 0.498, 0.702}
\definecolor{HighlightColor8}{rgb}{0.745, 0.769, 0.349}

\usepackage{tikz}
\usepackage{pgfplots}
\pgfplotsset{compat=1.16}

\usepackage[pdftex,colorlinks,linkcolor=HighlightColor,citecolor=HighlightColor,filecolor=HighlightColor,urlcolor=HighlightColor,backref,pdfpagelabels,pdfpagemode=UseNone,implicit]{hyperref}
\usepackage{comment}
\usepackage[inline]{enumitem}
\usepackage{listings}
\usepackage{caption}
\usepackage{subcaption}

\begin{document}
\pagestyle{empty}
\title{Compressing 1D Time-Channel Separable Convolutions \\using Sparse Random Ternary Matrices}
\name{Gon\c{c}alo Mordido$^{1*}$\thanks{*Work done during a research internship at NVIDIA.}, Matthijs Van keirsbilck$^2$, and Alexander Keller$^2$}
\address{ 
  $^1$Hasso Plattner Institute, Germany\\
  $^2$NVIDIA, Germany}
\email{goncalo.mordido@hpi.de, matthijsv@nvidia.com, akeller@nvidia.com}
\author{}
\maketitle

\begin{abstract}
We demonstrate that 1x1-convolutions in 1D time-channel separable convolutions may be replaced by
constant, sparse random ternary matrices with weights in $\{-1,0,+1\}$.
Such layers do not perform any multiplications and do not require training.
Moreover, the matrices may be generated on the chip during computation and
therefore do not require any memory access.
With the same parameter budget, we can afford deeper and more expressive models, improving the Pareto frontiers of existing models on several tasks. For command recognition on Google Speech Commands v1, we improve the state-of-the-art accuracy from 97.21\% to 97.41\% at the same network size.
Alternatively, we can lower the cost of existing models. For speech recognition on Librispeech, we half the number of weights to be trained while only sacrificing about $1\%$ of the floating-point baseline's word error rate.
\end{abstract}

\section{Introduction}

Speech and command recognition tasks tend to have strict low-latency requirements, which may be challenging to meet while using deep neural networks. Specifically, the employment of such networks on edge devices, which are bandwidth, energy, and area constrained, is often too costly in practical applications. Sparsity and quantization techniques are viable solutions, reducing the model size and computational cost while being suited for existing hardware. %

With this motivation in mind, we combine sparsity and quantization to make compact speech and command recognition models even more compact. In particular, we leverage recent residual networks~\cite{he2016deep} containing time-channel separable convolutions~\cite{QuartzNet,MatchboxNet,koluguri2020speakernet}. Despite such compact designs, most weights are on the 1x1-convolutions in the residual paths, which are equivalent to fully-connected layers applied to every time step. Hence, we propose to decrease the memory and computational cost by replacing these trained floating-point weights by constant, sparse random ternary matrices. %

Our approach enables
\begin{enumerate*}[label=(\roman*)]
    \item \textit{faster training}, since the weights need not be updated,
    \item \textit{faster on-chip inference}, due to sparse, low-precision operations with no multiplications, and
    \item \textit{higher compression rates}, because weights may not be stored but just computed on-the-fly resulting in
    \item \textit{reduced memory bandwidth}.
\end{enumerate*}

At a similar network size, we outperform several floating-point baselines on multiple data sets and tasks. We improve the current state-of-the-art on speech command recognition from 97.21\% to 97.41\% using MatchboxNet~\cite{MatchboxNet} on Google Speech Commands~\cite{google_speech_commands}. On AN4~\cite{an4}, we improve the Pareto efficiency in several floating-point configurations of QuartzNet~\cite{QuartzNet}, 
outperforming deeper and wider baselines at only  $\approx25\%$ and $\approx65\%$ of the network size, respectively. On Librispeech~\cite{librispeech}, when reducing the model size by half, we lose minimal performance, achieving 4.93\% WER as compared to the 3.90\% WER baseline.

\section{Related work}

After the advances of neural networks in the image domain, they have recently gained popularity in speech recognition.
Specifically, recurrent~\cite{graves2013hybrid} and convolutional \cite{abdel2014convolutional} neural networks, as well as a combination of both~\cite{sainath2015convolutional, amodei2016deep}, quickly have become common.
Encoder-decoder attention networks have also been proposed~\cite{bahdanau2016end}, and, following their adoption in the language domain, Transformers have been used for speech tasks as well~\cite{dong2018speech}.
However, recurrent networks are ill-suited to current parallel hardware due to their sequential nature, and attention mechanisms have a high computation and memory footprint.

Recent works show that convolutional networks may outperform recurrent networks even on sequence tasks \cite{bai2018empirical} while being better suited to modern parallel hardware. There have been many advances in the design of efficient convolutional models, especially in the computer vision community \cite{sandler2018mobilenetv2, tan2019efficientnet}. Similar network designs have been adapted to speech tasks, replacing 2D-convolutions by 1D-convolutions, \textit{e.g.} in Jasper~\cite{Jasper}, a residual convolutional network~\cite{he2016deep}.

Depth-wise separable convolutions have fewer parameters than standard convolutions while improving generalization performance~\cite{sandler2018mobilenetv2,hayashi2019einconv}.
Directly applying these lightweight convolutions to Jasper results in QuartzNet~\cite{QuartzNet}, which reduces the model size by $17\times$ at similar accuracy.
QuartzNet's residual blocks consist of time-channel separable convolutions, \textit{i.e.} across the time dimensions (temporal-convolutions) and across the feature dimensions (1x1-convolutions), followed by batch normalization (BN) and rectified linear units (ReLU).
MatchboxNet~\cite{MatchboxNet} adopts this design to achieve state-of-the-art results on some speech command recognition data sets. Similar models have also been proposed for speaker recognition and verification~\cite{koluguri2020speakernet} and cross-language learning and domain adaption~\cite{huang2020crosslanguage}.

On top of compact architectural design, other compression techniques may be used to further improve model efficiency. 
While pruning~\cite{molchanov2016pruning} attempts to find and remove unimportant weights, neurons, or layers, factorization~\cite{povey2018semi} compresses the weights using a low-dimensional approximation, and quantization~\cite{InstantANNQuantization, GradientQuantization} allows one to store and perform operations in low-bit precision reducing the computational cost significantly~\cite{han2015deep}. Considering ternary weights, \textit{e.g.} $\{-1, 0, 1\}$, multiplications are reduced to bit-wise, sparse operations in hardware, which are far less costly in terms of energy, speed, and area \cite{ttq}.

In another line of work, it has been shown that neural networks with random weights, \textit{e.g.} echo state networks~\cite{jaeger2001echo} and extreme learning machines~\cite{huang2006extreme}, can have surprisingly powerful capabilities~\cite{gallicchio2020deep}. These findings have also been shown to translate to automatic speech recognition tasks~\cite{skowronski2007automatic}.

In this work, we combine ternary quantization with random weights by replacing weights in 1x1-convolutions with constant, random ternary weights. 
As these weights neither need to be stored nor updated, a significant reduction in the memory and compute costs is possible in both training and inference time.

\section{Random ternary matrices}

\begin{figure}[h]
  \centering
  \includegraphics[width=0.4\linewidth]{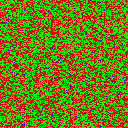} \qquad
  \includegraphics[width=0.4\linewidth]{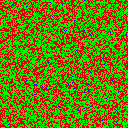}
  \caption{A trained ternary quantized weight matrix of a 1x1-convolution \cite{ttq} (left)
  can hardly be distinguished from a randomly generated ternary matrix (right).}
  \label{fig:TernaryMatrices}
\end{figure}

By analyzing the weights of the 1x1-convolutions in the sequence of
residual blocks, we observe that they do not expose any obvious visual structure.
Specifically, after applying trained ternarized quantization (TTQ~\cite{ttq}),
the resulting trained matrices maintain model performance even though they can hardly be distinguished from random ternary
matrices (see Figure~\ref{fig:TernaryMatrices}).

As previously mentioned, neural networks with random weights show powerful capabilities~\cite{jaeger2001echo,huang2006extreme,skowronski2007automatic}.
Therefore, we propose to use random ternarized matrices for the 1x1-convolutions in the residual blocks by keeping them constant while training the rest of the network.
For each neuron, the ternarized weights simply compute a scaled difference of sums
\begin{equation}
    \text{ReLU}\left(\text{BN}\left(\sum_{i \in I^+} x_i - \sum_{i \in I^-} x_i\right)\right) \, ,
\end{equation}
where the set $I^+$ contains all the indices of the connections with the weight $+1$
and $I^-$ contains all the indices of the connections with the weight $-1$.
The union $I:=I^- \cup I^+$ may be sparse, \textit{i.e.} may not contain the indices of all neurons in the previous layer. This scaled sum is a far cheaper operation than the traditional floating-point matrix multiplication.

Similar to echo state networks~\cite{jaeger2001echo}, a sufficient routing capacity is enabled by having trained layers before and after the constant random ternary layers. 
This matches observations in \cite{ttq,InstantANNQuantization}, where starting quantization too early in the network degrades performance.
With this in mind, we only ternarize and do not update the weights in the 1x1-convolutions inside the residual blocks (which possess the majority of the weights). Since the weights of the preceding temporal-convolution and the subsequent batch normalization layers are still updated, a scaling of the inputs and outputs may be learned, respectively.

Implicitly, the tensor product of the aforementioned scaling vectors defines the weights of the ternary matrix, empowering our approach.
We note that the ternary matrix in combination with batch normalization is reminiscent of
\textit{operational amplifiers}, computing a difference of sums that is
amplified before applying a non-linearity, \textit{e.g.} ReLU.
Moreover, our algorithm is related to extreme learning machines~\cite{huang2006extreme}, where a more theoretical foundation can be derived from \cite[Algorithm ELM]{ELM}.

\subsection{Implementation details}

Given a randomly initialized matrix, the random (and constant)
ternary matrix may be generated by applying the thresholding procedure
as used in \cite{ttq}. A parameter $t$ then serves as a threshold,
where all absolute values less than $t$ are quantized to zero and the
remaining values are quantized to -1 and +1 according to their sign.
If the matrices are initialized using uniformly distributed
samples from $[-1, 1]$, $t \in [0,1]$ will be the expected fraction of sparsity
of the matrix: $t = 0$ yields a dense, binary matrix, $t  = 1$ yields a
zero matrix, and $t \in ]0,1[$ yields a ternary matrix.
Listing~\ref{listing} shows a simple implementation of our approach using PyTorch.

\begin{lstlisting}[language=python,frame=single,caption=Example implementation given an uniformly distributed weight from $\mathrm{[-1, 1]}$\text{ and a threshold }$\mathrm{t \in [0,1]}$.,label=listing,morekeywords={int2}]
# compute signs
pos = (weight.data > 0.).int2()
neg = (weight.data < 0.).int2()

# compute mask
mask = (torch.abs(weight.data) > t).int2()

# make ternary
weight.data = (pos - neg) * mask

# do not update weight during training
weight.requires_grad = False
\end{lstlisting}

\subsection{Implementation variants}
Our approach is widely applicable and suited to existing hardware, allowing for a range of efficient implementations.
Specifically, the random ternary matrices may be computed on-the-fly on the chip without accessing memory for weights.
For example, a pseudo-random number generator may be used to determine the ternary values, where, given an initial state, a specific matrix may be deterministically generated at any time.
Alternatively, each ternary value may be determined by a hash function of the matrix entry coordinates, where different matrices may be realized by different hash functions.

Specific structured sparsity patterns supported by certain accelerators\footnote{https://developer.nvidia.com/blog/exploiting-ampere-structured-sparsity-with-cusparselt/} may be applied instead, taking full advantage of hardware support. Further compute acceleration in the ternary layers may be leveraged by custom layer implementations.

\section{Results}

We conducted experiments using MatchboxNet-NxMxW \cite{MatchboxNet} on Google Speech Commands (GSC)~\cite{google_speech_commands} and QuartzNet-NxMxW \cite{QuartzNet} on AN4~\cite{an4} for 200 epochs and Librispeech~\cite{librispeech} for 300 epochs. As for the notation, N stands for the number of residual blocks, M for the number of depth-wise separable convolutions in the residual blocks, and W for the number of channels. We refer to the original papers for additional architectural details.

GSCv1 consists of 65k one-second spoken utterances from various speakers. There are 30 command classes chosen for possible IoT or robotics application scenarios, numerical digits, as well as phoneme variety and command similarity. %
AN4 is an alphanumeric data set consisting of around 1k utterances averaging 3 seconds long, describing personal information and control words. Librispeech contains 10k hours of segmented and aligned narrations of public audiobooks and is commonly used to evaluate large-scale speech recognition models.

In our experiments, we compared our simplified networks to the original baselines under identical training scenarios. We adopted the training setups publicly available at NVIDIA NeMo\footnote{https://github.com/NVIDIA/NeMo}, so that our results may be easily replicated as well as applied to additional models and tasks.

\subsection{Speech command recognition (MatchboxNet on GSC)}

\begin{figure}[t]
  \centering
  \raisebox{0.2ex}{a.} \includegraphics[width=0.96\linewidth]{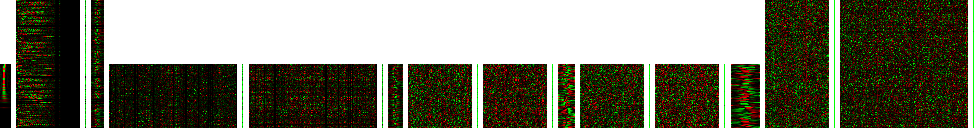}
  \raisebox{0.2ex}{b.} \includegraphics[width=0.96\linewidth]{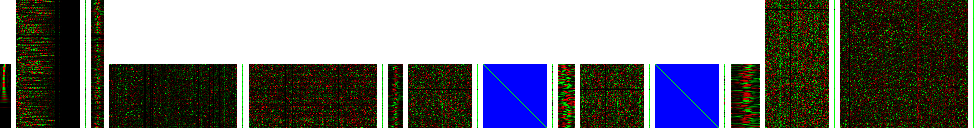}
  \raisebox{0.2ex}{c.} \includegraphics[width=0.96\linewidth]{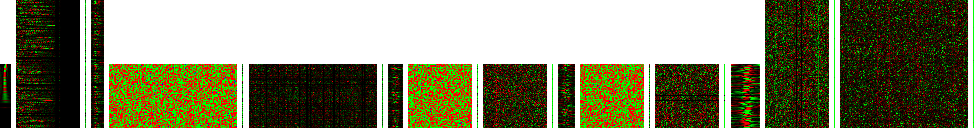}
  \raisebox{0.2ex}{d.} \includegraphics[width=0.96\linewidth]{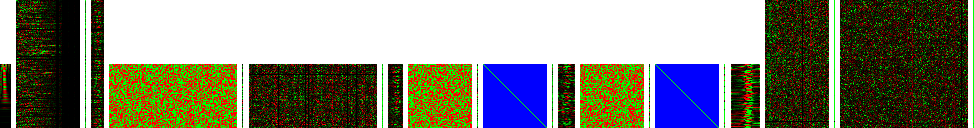}
  \raisebox{0.2ex}{e.} \includegraphics[width=0.96\linewidth]{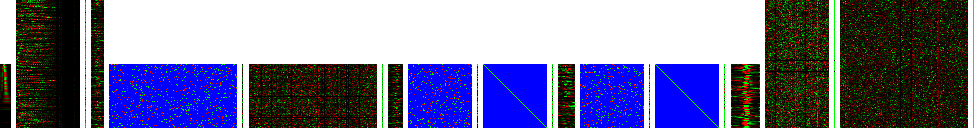}
  
  \caption{MatchboxNet-3x1x64 matrices: colors indicate negative weights (red) or positive weights (green). To illustrate sparsity, zero is shown in blue in constant matrices.
  Row a) shows the baseline's trained floating-point weights;
  b) uses identity skip links;
  c) replaces the 1x1-convolutions in residual layers by constant, random ternary matrices in which red is $-1$ and green is $+1$;
  d) combines b) and c); 
  e) uses sparse ternary weights.}
  \label{fig:MatchboxNet}
\end{figure}

\begin{figure}
  \centering
  \includegraphics[width=\linewidth]{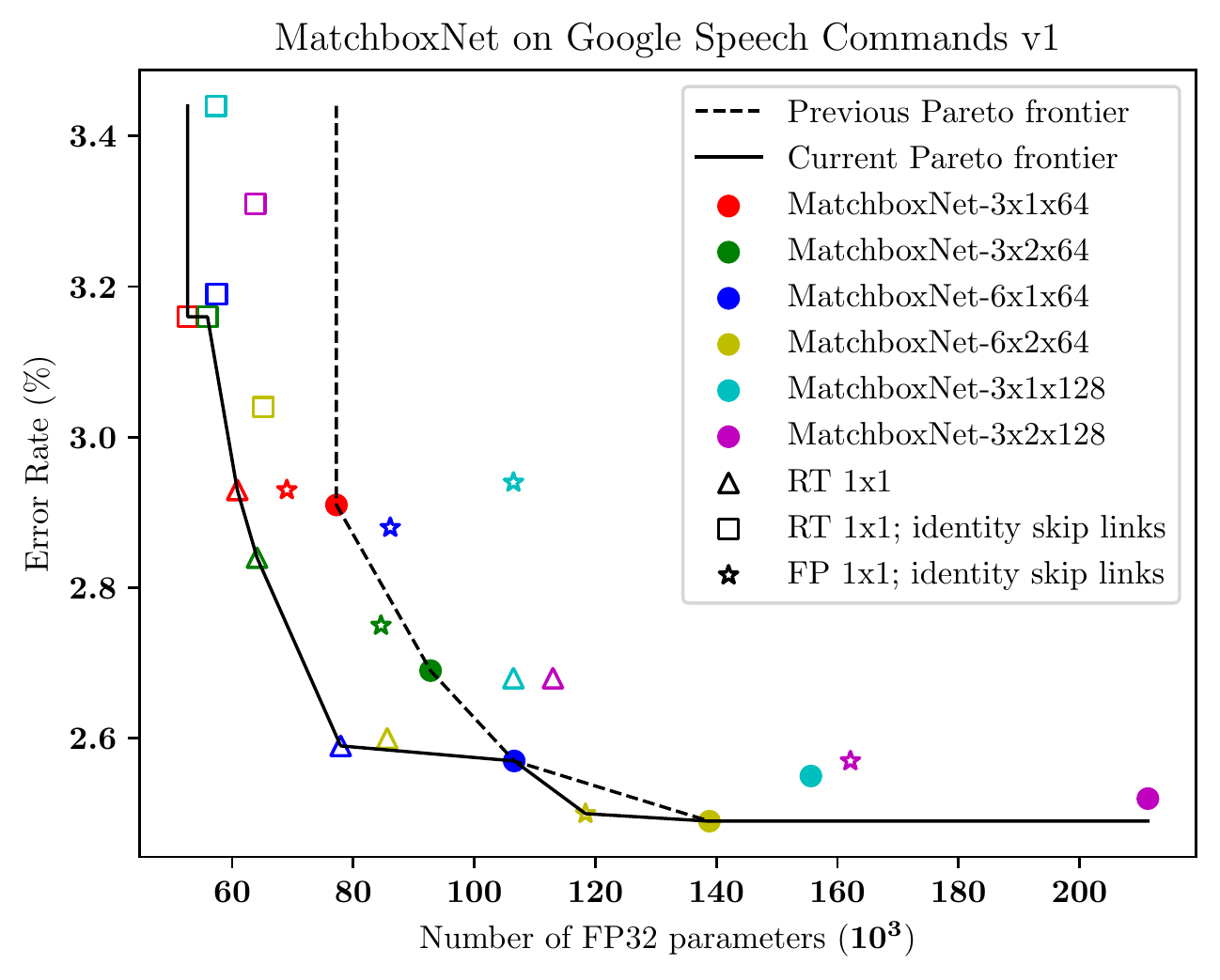}
  \caption{Test error rate of different MatchboxNet configurations
  with respect to the number of model parameters on GSCv1. 
  "FP" refers to floating-point and "RT" to random ternary weights. 
  Our constant, random ternary
  matrices improve the Pareto frontier. Restricting the skip links to identity matrices greatly reduces the network size but increases error.
  }
  \label{fig:MatchboxNetParameters}
\end{figure}

Figure~\ref{fig:MatchboxNet} shows the weight matrices resulting from multiple model size reductions. Figure~\ref{fig:MatchboxNetParameters} compares the test accuracy of several MatchboxNet configurations with respect to the number of model parameters. Training for 200 epochs, our MatchboxNet-3x1x64 baseline achieves 97.09\% test accuracy, which is close to the mean of 97.21\% accuracy over five trials reported in \cite{MatchboxNet}. 
We observe that using random ternary matrices improves the Pareto efficiency of the models, reaching a better trade-off of error rate for model size and cost.
Hence, using random ternary weights with and without identity skip links allows us to increase model width and depth efficiently. 
We trained our models with varied $t$ values and plot the best models for each configuration. Note that the sparsity levels $t$ have little influence on the overall performance as discussed later in Section \ref{sec:an4}.

Although ternarizing the skip links reduces performance in our experiments, using identity skip links avoids the otherwise quadratic weight matrices and performs close to the baseline for the smaller model configurations. The skip links even may be omitted. Yet deeper models train better with skip links~\cite{he2016deep,he2016identity}.
Interestingly, tests using constant floating-point random matrices did not result in any improvements as compared to our more efficient random ternary matrices.

Table \ref{tab:results_matchboxnet} compares our random ternary variants to the floating-point models reported in~\cite{MatchboxNet}. At the same model size (77K), our random ternary MatchboxNet-6x1x64 sets a new state-of-the-art of 97.41\% accuracy. Our smaller variants further reduce model size while still maintaining high accuracy.

\begin{table}[h]
  \caption{Accuracy and model size on Google Speech Commands version 1. "FP" denotes floating-point weights and "RT" denotes our constant, random ternary weights.}
  \label{tab:results_matchboxnet}
  \centering
  \begin{tabular}{ccc}
    \toprule
    \textbf{Model} & \textbf{FP32} & \textbf{Acc.} \\
     & \textbf{param. (K)} & \textbf{(\%)} \\
    \midrule
    DenseNet-BC-100~\cite{li2020feature} & 800 & 96.77\\
    ResNet-15~\cite{DBLP:conf/interspeech/TangSDM18} & 238 & 95.80\\
    EdgeSpeechNet-A~\cite{lin2018edgespeechnets} & 107 & 96.80\\
    MatchboxNet-3x1x64~\cite{MatchboxNet} (FP) & 77 & 97.21\\
    \midrule
    MatchboxNet-6x1x64 (RT) & 77 & 97.41\\
    MatchboxNet-3x2x64 (RT) & 64 & 97.16\\
    MatchboxNet-3x1x64 (RT) & 60 & 97.07\\
    \bottomrule
  \end{tabular}
\end{table}

\subsection{Automatic speech recognition}

In Section~\ref{sec:an4}, we use the fast iterations enabled by using the small AN4 data set to study a range of sparsity levels as well as the Pareto efficiency on multiple QuartzNet configurations.
We then extrapolate these findings and evaluate our method against large-scale QuartzNet models on Lirispeech (Section~\ref{sec:librispeech}).

\subsubsection{QuartzNet on AN4}
\label{sec:an4}

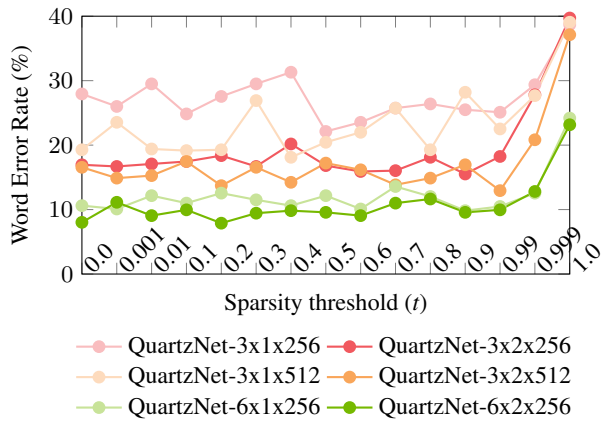
\begin{figure}[ht]
  \centering
\begin{tikzpicture} %
  \begin{axis}[%
      xlabel=Sparsity threshold (\textit{t}), ylabel=Word Error Rate (\%), 
      xmin=0, xmax=14, ymin=0, ymax=40, 
      height=5cm, width=\linewidth,
      xtick={0,1,2,3,4,5,6,7,8,9,10,11,12,13,14},
      xticklabels={0.0, 0.001, 0.01, 0.1, 0.2, 0.3, 0.4, 0.5, 0.6, 0.7, 0.8, 0.9, 0.99, 0.999, 1.0},
      x tick label style = {rotate=45, anchor=west},
      legend style={draw=none, at={(0,0)}, anchor=north, at={(axis description cs:0.5,-0.2)}},
      legend columns=2
      ]

      \addplot[thick,color=HighlightColor2!40,mark=*] coordinates {
      (0,27.94) (1,26.0) (2,29.5) (3,24.84) (4,27.55) (5,29.5) (6,31.31) (7,22.12) (8,23.54) (9,25.74) (10,26.39) (11,25.49) (12,25.1) (13,29.37) (14,38.68)};
      \addlegendentry{QuartzNet-3x1x256}
      \addplot[thick,color=HighlightColor2,mark=*] coordinates {
      (0,16.95) (1,16.69) (2,17.08) (3,17.46) (4,18.37) (5,16.69) (6,20.18) (7,16.82) (8,15.91) (9,16.04) (10,18.11) (11,15.52) (12,18.24) (13,27.81) (14,39.72)};
      \addlegendentry{QuartzNet-3x2x256}

      \addplot[thick,color=HighlightColor4!40,mark=*] coordinates {
      (0,19.28) (1,23.54) (2,19.4) (3,19.15) (4,19.28) (5,26.91) (6,18.11) (7,20.44) (8,21.99) (9,25.74) (10,19.28) (11,28.2) (12,22.51) (13,27.68) (14,39.07)};
      \addlegendentry{QuartzNet-3x1x512}
      \addplot[thick,color=HighlightColor4,mark=*] coordinates {
      (0,16.56) (1,14.88) (2,15.27) (3,17.46) (4,13.71) (5,16.56) (6,14.23) (7,17.21) (8,16.17) (9,13.84) (10,14.88) (11,16.95) (12,12.94) (13,20.83) (14,37.13)};
      \addlegendentry{QuartzNet-3x2x512}

      \addplot[thick,color=HighlightColor1!40,mark=*] coordinates {
      (0,10.61) (1,10.09) (2,12.16) (3,11.0) (4,12.55) (5,11.51) (6,10.61) (7,12.16) (8,10.09) (9,13.58) (10,12.03) (11,9.83) (12,10.48) (13,12.55) (14,24.19)};
      \addlegendentry{QuartzNet-6x1x256}
      \addplot[thick,color=HighlightColor1,mark=*] coordinates {
      (0,8.02) (1,11.13) (2,9.06) (3,9.96) (4,7.89) (5,9.44) (6,9.83) (7,9.57) (8,9.06) (9,11.0) (10,11.64) (11,9.57) (12,9.96) (13,12.81) (14,23.16)};
      \addlegendentry{QuartzNet-6x2x256}
  \end{axis}
\end{tikzpicture}
  \caption{Word error rate of different QuartzNet configurations using our method on the AN4 data set.
  High sparsity levels ($t$ close to 1) deliver performance similar to dense weights ($t$ close to 0). Bigger networks are less sensitive to different thresholds.}
  \label{fig:sparsity_levels}
\end{figure}

Figure~\ref{fig:sparsity_levels} shows that at 90\% average sparsity, i.e. $t = 0.9$, different configurations of QuartzNet perform comparably to the less computationally efficient fully dense models. We find a similar pattern for MatchboxNet. Therefore, we can choose a highly sparse network to minimize memory and compute cost. The performance starts deteriorating as $t \rightarrow 1$ as then the 1x1-convolutions get too sparse, eventually becoming zero matrices.

\begin{figure}[ht]
  \centering
  \includegraphics[width=\linewidth]{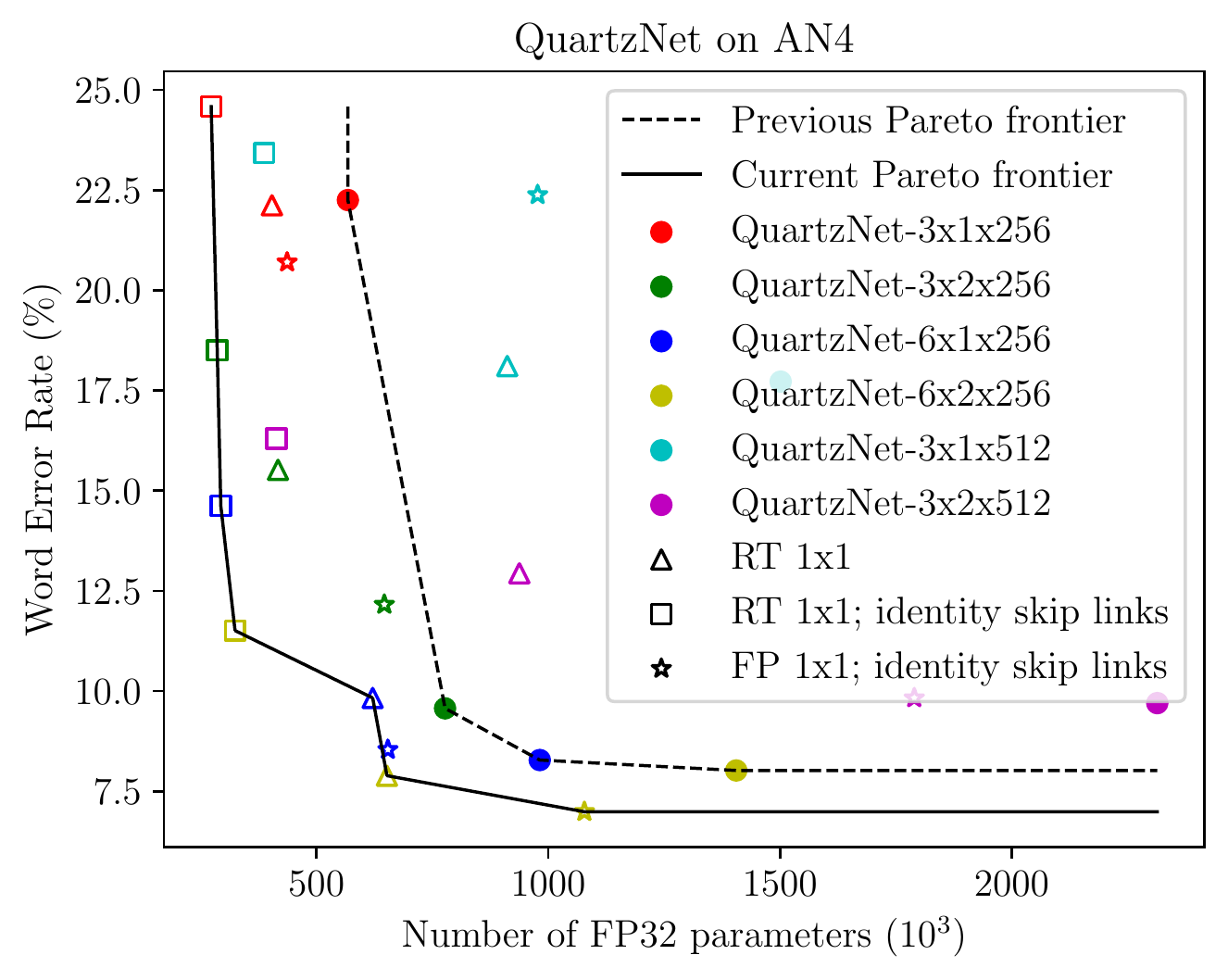}
  \caption{Test word error rate of different QuartzNet configurations with respect to the number of model parameters on AN4.
      "FP" refers to floating-point weights and "RT" to constant, random ternary weights.
      The best Pareto front is formed by our random ternary variants, with and without identity skip links.}
  \label{fig:QuartzNetParameters}
\end{figure}

Figure~\ref{fig:QuartzNetParameters} compares several QuartzNet configurations in terms of word error rate and network size.
Our constant, random ternary variants, with and without identity skip links, create a new Pareto frontier, improving the performance of smaller models.

\subsubsection{QuartzNet on Librispeech}
\label{sec:librispeech}

Using the original QuartzNet-15x5~\cite{QuartzNet}, we randomize and ternarize the 1x1-convolutions in the latter residual blocks, as those layers contain the most parameters. 
Out of the 15 blocks, we ternarize the last 6 blocks, reducing the network size by half. %
We used trained skip links and $t = 0.001$ as our threshold, although an iteration over multiple $t$ values, similarly to our previous experiments, would likely lead to better results.

Table~\ref{tab:results_librispeech_test} compares the greedy WER of our variant to the bigger models as reported in~\cite{QuartzNet}. We observe that, despite the significant reduction of the number of floating-point parameters, our variant achieves competitive results: using half the parameters costs only $\approx1\%$ WER as compared to the floating-point QuartzNet-15x5.

We compare our variant to shallower QuartzNet configurations reported in~\cite{QuartzNet} in Table~\ref{tab:results_librispeech_dev}. %
Extrapolating the performance of such configurations at the same network size as our variant, we see that random ternary weights either maintain or increase parameter efficiency on dev-clean and dev-other, respectively.

\begin{table}[ht]
  \caption{Word error rate and model size on Librispeech. "FP" denotes floating-point weights and "RT" denotes our random, ternary weights.}
  \label{tab:results_librispeech_test}
  \centering
  \begin{tabular}{cccc}
    \toprule
    \textbf{Model} & \textbf{FP32} & \multicolumn{2}{c}{\textbf{Test}}\\
     & \textbf{param. (M)} & \textbf{clean} & \textbf{other}\\
    \midrule
    JasperDR-10x5~\cite{Jasper} & 333 &  4.32 & 11.82\\
    TDS Conv~\cite{Hannun2019} & 37 & 5.36 & 15.64\\
    QuartzNet-15x5~\cite{QuartzNet} (FP) & 19 & 3.90 & 11.28\\
    \midrule
    QuartzNet-15x5 (RT) & 9 & 4.93 & 14.21\\
    \bottomrule
  \end{tabular}
\end{table}

\begin{table}[ht]
  \caption{Word error rate and model size of several QuartzNet variants on Librispeech. "FP" means floating-point weights and "RT" denotes our random, ternary weights. Underlined results are extrapolated from the Pareto frontier of the floating-point baselines to match the model size of our simplified variant.%
  }
  \label{tab:results_librispeech_dev}
  \centering
  \begin{tabular}{cccc}
    \toprule
    \textbf{Model} & \textbf{FP32} & \multicolumn{2}{c}{\textbf{Dev}}\\
     & \textbf{param. (M)} & \textbf{clean} & \textbf{other}\\
    \midrule
    QuartzNet-15x5~\cite{QuartzNet} (FP) & 19 & 3.98 & 11.58\\
    QuartzNet-10x5~\cite{QuartzNet} (FP) & 13 & 4.14 & 12.33\\
    QuartzNet (FP, extrap.) & 9 & \underline{4.84} & \underline{14.20}\\
    QuartzNet-5x5~\cite{QuartzNet} (FP) & 7 & 5.39 & 15.69\\
    \midrule
    QuartzNet-15x5 (RT) & 9 & 4.84 & 13.75\\
    \bottomrule
  \end{tabular}
\end{table}

\section{Conclusion and future work}

Random ternary 1x1-convolutions 
improve the efficiency of speech residual networks, reducing memory and computational cost significantly during training and inference. Our method is energy efficient and simple to support in hardware:
while computation may be realized on-chip without multiplications,
 memory access may be avoided by computing the weights on-the-fly.
Hence, our random ternary layers are a simple, yet effective way to further compact existing network designs.%

Due to its similarity to operational amplifiers, our algorithm may be explored in analog hardware in the future. %

\section{Acknowledgements}
The third author is very thankful to C\'edric Villani for a discussion
on structure to be discovered in neural networks
during the AI for Good Global Summit 2019 in Geneva.
This work has been partially funded by the Federal Ministry of Education and Research (BMBF, Germany) in the project Open Testbed Berlin - 5G and Beyond - OTB-5G+ (F\"orderkennzeichen 16KIS0980).

\clearpage

\end{document}